\documentclass[lettersize,journal]{IEEEtran}
\usepackage{amsmath,amsfonts}
\usepackage{algorithmic}
\usepackage{algorithm}
\usepackage{array}
\usepackage[caption=false,font=normalsize,labelfont=sf,textfont=sf]{subfig}
\usepackage{textcomp}
\usepackage{stfloats}
\usepackage{url}
\usepackage{verbatim}
\usepackage{graphicx}
\usepackage{cite}
\usepackage{times}
\usepackage{epsfig}
\usepackage{graphicx}
\usepackage{amsmath}
\usepackage{amssymb}
\usepackage{bm}
\usepackage{xcolor}
\begin{document}

\title{Dense Supervision Propagation for Weakly Supervised Semantic Segmentation on 3D Point Clouds}

\author{Jiacheng Wei, Guosheng Lin,
        Kim-Hui Yap, Fayao Liu, and Tzu-Yi Hung%
        % <-this % stops a space

\thanks{Manuscript received June 26, 2023; revised September 29, 2023, October 26, 2023, and November 6, 2023. \textit{(Corresponding author: Guosheng Lin.)}}
\thanks{Jiacheng Wei and Kim-Hui Yap are with the School of Electrical and Electronic Engineering, Nanyang Technological University, Singapore 639798.
(E-mail: jiacheng002@e.ntu.edu.sg, ekhyap@ntu.edu.sg) }
\thanks{Guosheng Lin is with the School of Computer Science and Engineering, Nanyang Technological University, Singapore 639798. (e-mail:
gslin@ntu.edu.sg).}
%Digital Object Ident}
% \IEEEcompsocthanksitem Wei J., Wang H. and Lin G. are with Nanyang Technological University.\protect\\
\thanks{Fayao Liu is with Institute for Infocomm Research A*STAR, Singapore.}% <-this % stops a space
\thanks{Tzu-Yi Hung is with the Delta Research Center, Singapore.}}

\markboth{IEEE TRANSACTIONS ON CIRCUITS AND SYSTEMS FOR VIDEO TECHNOLOGY}%
{}

\IEEEpubid{\begin{minipage}{\textwidth}\ \\[24pt] \centering
  Copyright © 2023 IEEE. Personal use of this material is permitted. \\
  However, permission to use this material for any other purposes must be obtained from the IEEE by sending an email to pubs-permissions@ieee.org.
\end{minipage}}


\maketitle

\begin{abstract}
Semantic segmentation on 3D point clouds is an important task for 3D scene understanding. While dense labeling on 3D data is expensive and time-consuming, only a few works address weakly supervised semantic point cloud segmentation methods to relieve the labeling cost by learning from simpler and cheaper labels. Meanwhile, there are still huge performance gaps between existing weakly supervised methods and state-of-the-art fully supervised methods. %\textcolor{blue}{
In this paper, we propose Dense Supervision Propagation (DSP) to train a semantic point cloud segmentation network with only a small portion of points being labeled. We argue that we can better utilize the limited supervision information as we densely propagate the supervision signal from the labeled points to other points within and across the input samples. Specifically, we propose a cross-sample feature reallocating module to transfer similar features and therefore re-route the gradients across two samples with common classes and an intra-sample feature redistribution module to propagate supervision signals on unlabeled points across and within point cloud samples. We conduct extensive experiments on public datasets S3DIS and ScanNet. Our weakly supervised method with only 10\% and 1\% of labels can produce competitive results with the fully supervised counterpart.
\end{abstract}


\begin{figure}[t]
\begin{center}
   \includegraphics[width=0.99\linewidth]{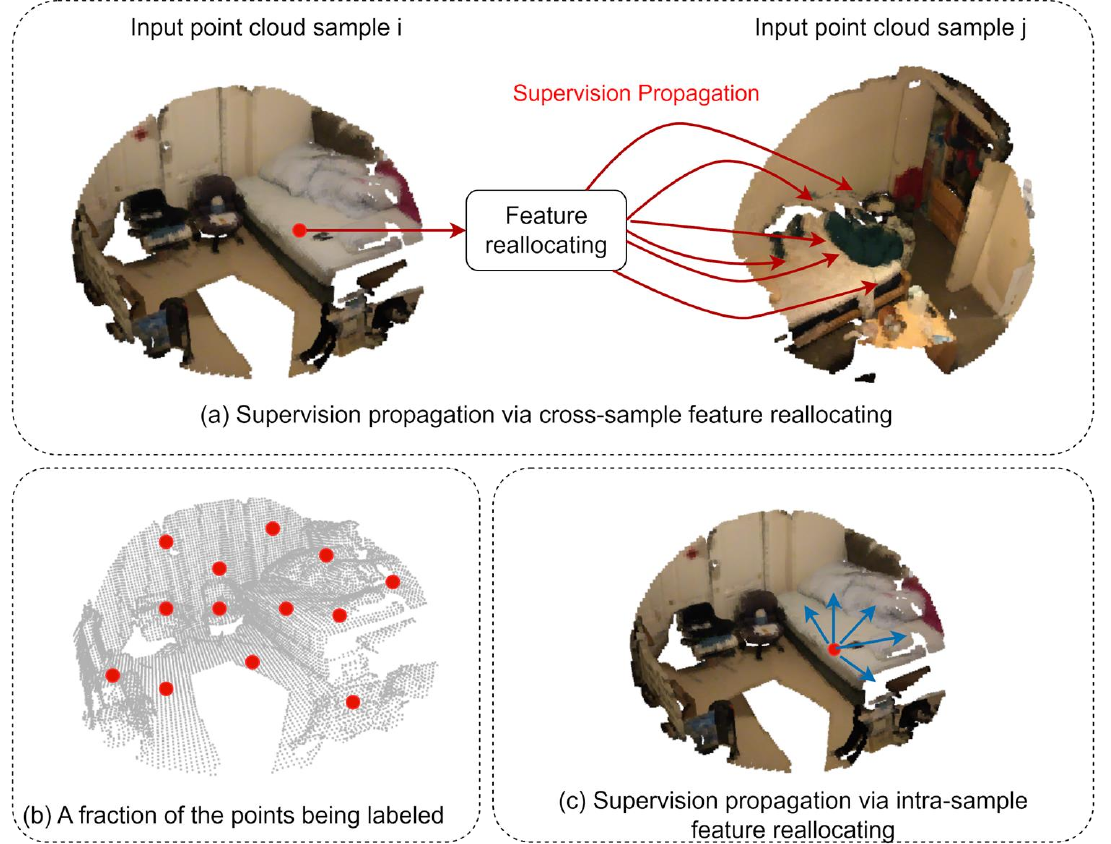}
\end{center}
   \caption{(a) Supervision propagation across samples via cross-sample feature reallocating. For clarity, this figure shows only a single direction of the feature reallocating and supervision propagation, the module can propagate supervision bidirectionally. (b) An illustration of our weak labels. (c) Supervision propagation within samples via intra-sample feature reallocating.}
%\label{fig:long1}
\label{fig:1}
%\vspace{-10pt}
\end{figure}

\section{Introduction}
\IEEEPARstart{W}{ith} the development of 3D sensors, and point cloud data are playing important roles in multimedia applications. Semantic point cloud segmentation is vital for 3D scene understanding, which provides fundamental information for further applications like augmented reality and mobile robots.
Recent developments in deep learning-based point cloud analysis methods have made considerable progress in 3D semantic segmentation\cite{han2020occuseg,choy20194d,graham20183d,thomas2019kpconv}. However, 3D semantic segmentation requires point-level labels, which are much more expensive and time-consuming than the labels of 3D classification and detection tasks. 

%change
%\textcolor{blue}{
Recently, in the realm of circuits and systems for video technology, many efforts have been put into weakly supervised semantic segmentation (WSSS)\cite{8941066, 9612214, 10015839} on 2D images and achieved remarkable results. However, despite dense labeling on point clouds being even more expensive than dense image labeling, only a few works have been done on 3D WSSS\cite{su2023weakly, 10035004} and yet there remains a huge performance gap with the state-of-the-art fully supervised methods. 
%}
%change

The existing 3D WSSS methods formulate the problem in different directions. \cite{wang2020weakly} utilize dense 2D segmentation labels to supervise the training in 3D by projecting the 3D predictions onto the corresponding 2D labels. However, each 3D sample is projected to 2D in several views and each projected 2D image needs pixel-level labeling. Thus, this method still requires a large amount of manual labeling. \cite{wei2020multi} proposes to generate pseudo point-level label using 3D class activation map\cite{zhou2016learning} from subcloud-level annotation, which is similar to the 2D WSSS methods using image-level labels. \cite{xu2020weakly} directly trains a point cloud segmentation network with 10 times fewer labels, which is close to point supervision\cite{bearman2016s} and scribble supervision\cite{lin2016scribblesup, wang2019boundary} in 2D WSSS methods. We adopt the 3D WSSS setting from \cite{xu2020weakly} that takes only a small fraction of points to be labeled. 
%\textcolor{blue}{
In this study, our primary objective is to optimize the use of scarce annotations by intensifying the dense propagation of supervision signals from labeled to unlabeled points. Initially, we employ the cutting-edge point cloud segmentation network, KPConv\cite{thomas2019kpconv}, as our foundational model. To overcome the aforementioned challenges, we introduce a novel two-stage training strategy, incorporating both our cross-sample feature reallocating module (CSFR) and the intra-sample feature reallocating (ISFR) module.%}

%\textcolor{blue}{
As depicted in Figure\ref{fig:1}, the first stage of our training process draws inspiration from \cite{sun2020mining, liu2020weakly, fan2020cian}. Here, we select two samples with at least one overlapping class to serve as an input pair. The CSFR module is designed to facilitate the transfer of analogous features between these two samples. Unlike methods in \cite{sun2020mining, liu2020weakly, fan2020cian} which directly incorporate warped features from one sample to detect more active regions, we opt to reconstruct the features using the point correlation of the data pair, instead of mere addition. Consequently, each point in the restructured feature emerges as a weighted aggregate of all points from its counterpart sample. We then deploy a cross-regularization loss to gently enforce weak supervision upon these reallocated features. This approach enables us to effectively channel the gradients, allowing for a dense propagation of supervision from labeled points in one sample to their unlabeled counterparts in the other.%}

%\textcolor{blue}{
In the subsequent training stage, our ISFR module facilitates the transmission of supervision from labeled to unlabeled points within each individual sample, once again utilizing feature reallocation based on point correlation. This ensures that supervision is densely transmitted from labeled to unlabeled points within a given sample. Given the potential for non-overlapping classes between input pairs, which could introduce noise during the supervision propagation, our strategy involves training the network using the CSFR module to first acquire general representations. This is followed by employing the ISFR module for fine-tuning in the latter stage. As both modules function based on point correlations, supervision signals are effectively propagated to unlabeled points bearing resemblance in features to labeled ones. It is imperative to note that the primary role of these two modules is to enhance the training of the basic network. They are redundant for generating the final segmentation predictions during the inference phase. These modules indirectly steer the training of the basic segmentation network and remain unused during testing.

Compared to MulPro\cite{su2023weakly} and MP\cite{10035004} which focus on exploring the information within the current input sample, our methods leverage to explore the information that contains inter-sample to find more supervision cues for weakly supervised 3D semantic segmentation task.

In summary, our contributions are:
\begin{itemize}
    \item We propose a cross-sample feature reallocating module to reconstruct features and re-route gradients across the input pair based on point correlation. Hence, the supervision signals from labeled points can be propagated to unlabeled points across samples.
    \item We propose an intra-sample feature reallocating module to reallocate features within each sample based on point correlation. Then, the supervision signals can be propagated from labeled points to unlabeled points within each point cloud sample.
    \item We propose a two-stage training strategy so the cross-sample feature reallocating module and the intra-sample feature reallocating module can both contribute to the performance without interfering with each other. 
    \item Our weakly supervised methods with only 10\% and 1\% of the points being labeled can produce compatible results with their fully supervised counterpart in S3DIS and ScanNet datasets.
\end{itemize}

%-------------------------------------------------------------------------
\begin{figure*}[t]
\begin{center}
%\fbox{\rule{0pt}{2in} \rule{.9\linewidth}{0pt}}
\includegraphics[width=0.99\linewidth]{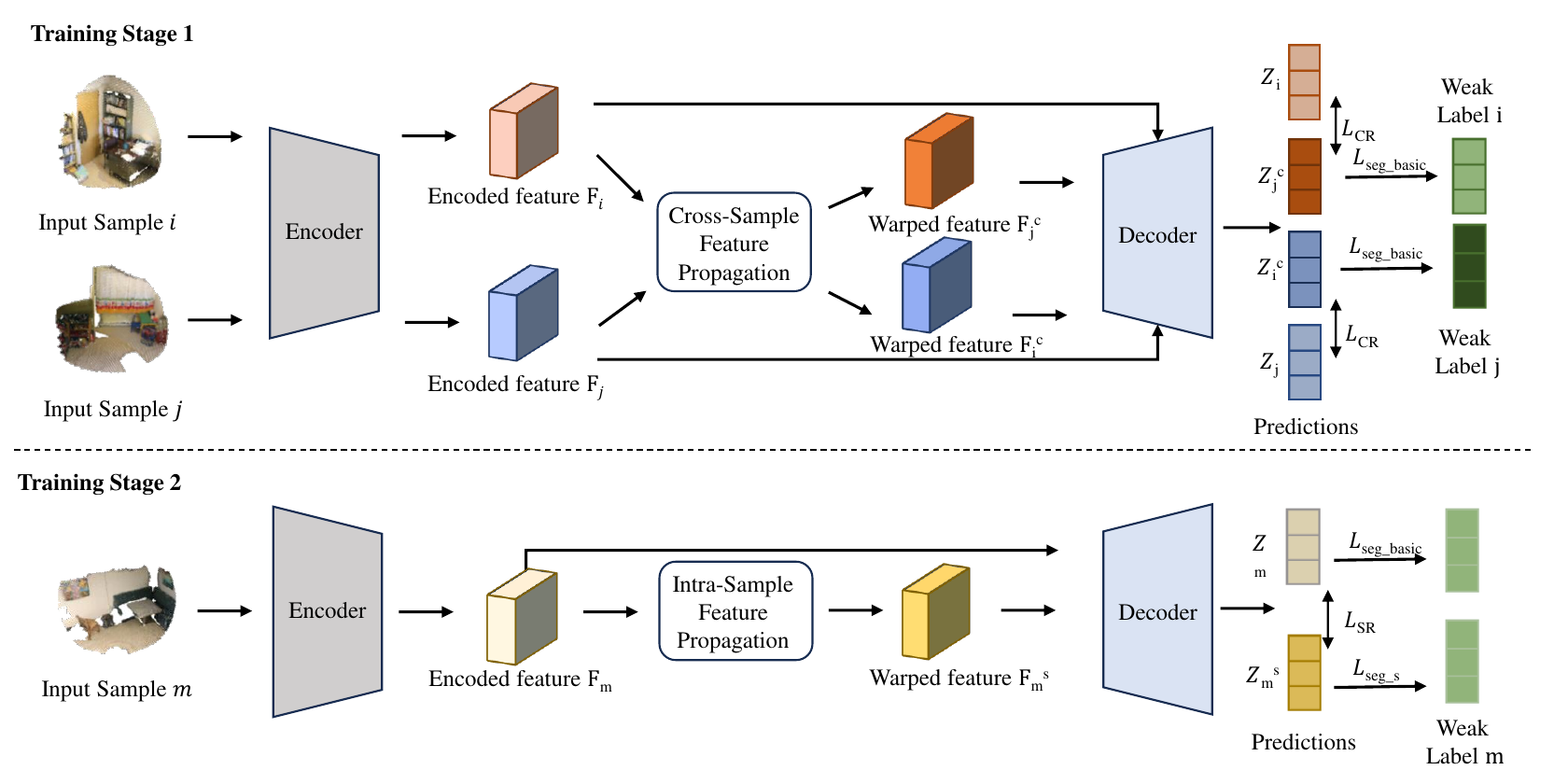}
\end{center}
   \caption{
   The overall two-stage training framework for our proposed method. In Stage 1, a pair of point cloud samples are input into the encoder, yielding encoded features $F_i$ and $F_j$. A bespoke cross-sample feature reallocation module is employed to redistribute these features and direct gradient flows, thereby facilitating the propagation of supervisory signals between the pair. This results in the formation of transformed features $F_j^c$ and $F_i^c$. Subsequently, in Stage 2, individual point cloud samples are independently processed by the network to produce the encoded feature $F_m$. Here, an intra-sample feature propagation module is designed to internalize supervisory signals, obtaining the modified feature $F_m^s$. Throughout both stages, the encoded and warped features are conveyed to the decoder to generate the final predictions. It is important to note that the encoder and decoder are consistent across both stages, with the encoder and decoder parameters from Stage 1 being transferred to Stage 2.
}
\label{fig:framework}
\vspace{-10pt}
\end{figure*}
%-------------------------------------------------------------------------

\section{Related Works}
\subsection{Semantic Segmentation on 3D Point clouds}
There are three categories for 3D semantic segmentation methods: projection-based methods, voxel-based methods and point-based methods. Multi-view projection-based methods\cite{kundu2020virtual, zhang2020fusion, dai20183dmv} project the 3D data into 2D from multiple viewpoints, therefore they can easily process the projected data on 2D convolution networks. However, these methods suffer from occlusion, view-point selection, misalignment, and other defects that may limit the performance. Voxel-based methods like Submanifold Sparse Convolution\cite{graham20183d}, MinkowskiNet\cite{choy20194d} and Occuseg\cite{han2020occuseg} first quantize the point cloud data into voxels and perform sparse 3D convolution on the voxels. These methods can often achieve good segmentation performance but severely suffer from heavy memory and time consumption. Point-based methods directly take raw point cloud data as network input. PointNet\cite{qi2017pointnet} and PointNet++\cite{qi2017pointnet++} process point cloud data with stacked MLPs. DGCNN\cite{wang2019dynamic} proposes EdgeConv to capture local geometry by dynamically generating graphs for points with their neighbors. PointCNN\cite{li2018pointcnn} and PointConv\cite{wu2019pointconv} formulate convolution operations in 3D using KNNs for each point. KPConv\cite{thomas2019kpconv} uses filters and kernel points in Euclidean space to formulate convolution operations with distances of points within the filter.

%-------------------------------------------------------------------------
\subsection{Weakly supervised 2D semantic segmentation}
Most 2D WSSS methods use image-level labels. Based on class activation map(CAM)\cite{zhou2016learning}, many methods\cite{wang2020self,wei2018revisiting,huang2018weakly,shen2018bootstrapping,shen2018bootstrapping, ahn2018learning, kolesnikov2016seed,sun2020mining, liu2020weakly, fan2020cian} refines the CAM generated from a classification network to generate pseudo-pixel-level labels. Then, segmentation networks are trained using the pseudo-pixel-level labels. Besides the image-level label, other kinds of weak labels like point supervision\cite{bearman2016s} and scribble supervision\cite{lin2016scribblesup, wang2019boundary} which is similar to the weak setting in this work. Points and scribble supervision methods usually constrain the unlabeled points using label consistency with local smoothness.

%-------------------------------------------------------------------------
\subsection{Weakly supervised 3D semantic segmentation}
 Existing 3D WSSS methods utilize different kinds of weak supervisions.\cite{wang2020weakly} utilize dense 2D segmentation labels to supervise the training in 3D by projecting the 3D predictions onto the corresponding 2D labels. \cite{wei2020multi} proposes to generate pseudo point-level label using 3D class activation map\cite{zhou2016learning} from subcloud-level annotation. \cite{xu2020weakly} directly trains a point cloud segmentation network with 10 times fewer labels. 
 Scribble-Sup\cite{unal2022scribble} use scribble annotations on the Lidar data as supervision. MulPro\cite{su2023weakly} incorporated prototypes during the training to expand the weak labels. PSD\cite{zhang2021perturbed} proposed a Self-Distillation framework that leverages self-supervised learning principles. DAT\cite{wu2022dual} proposed dual adaptive transformations to learn localization cues from both point-level and region-level. WS3\cite{zhang2021weakly} proposed a pretext task as point colorization to explore the information contained in the data. WS3D\cite{liu2022weakly} focused on the region boundary with an energy-level loss to improve the segmentation results. Contrastive Scene Contexts\cite{hou2021exploring} explores contrastive self-supervised learning to explore cues within the training data. HybridCR\cite{li2022hybridcr} employs a contrastive loss function computed not only on the original sample but also on augmented samples derived from the original data. Their primary emphasis lies in evaluating the similarity between the original sample and its augmented counterparts. However, all these methods failed to explore the inter-sample information in the 3D point cloud data.

%-------------------------------------------------------------------------
\section{Approach}
As illustrated in Figure\ref{fig:framework}, we propose a two-stage training strategy. In the first training stage, we train the basic segmentation network with the cross-sample feature reallocating module. And in the second stage, we train the basic segmentation network with the intra-sample feature reallocating module. We will explain each module in detail.
%-------------------------------------------------------------------------
\subsection{Basic segmentation network}
The basic component of our method is a point cloud segmentation network. We take the 
state-of-the-art deep learning architecture Kernel Point Convolution(KPConv)\cite{thomas2019kpconv} segmentation network as our backbone network.
We denote each input point cloud sample as $\bm{P}_i \in \mathbb{R}^{N_i \times D}$ where $N_i$ is the input point number and $D$ is the input feature dimension. After feeding the $\mathit{P_i}$ to the encoder, we get the encoded feature map $\bm{F}_i=f_{enc}(\bm{P}_i)\in\mathbb{R}^{N^e_i \times K}$, where $N^e_i$ is the down-sampled point numbers for sample $i$ and $K$ is the feature dimension. The encoded feature $\bm{F}_i$ is then fed into the decoder to get the final prediction $\bm{Z}_i = f_{dec}(\bm{F}_i) \in \mathbb{R}^{N \times C}$, where $C$ is the number of classes. The one-hot label of the sample is denoted as $\bm{Y}_i \in \mathbb{R}^{N \times C}$. However, as we only have labels for a small fraction of the points, we define a binary mask $\bm{M} \in \mathbb{R}^N$ to indicate whether a point is being labeled, the mask is 1 for labeled points and 0 for unlabeled points. Then, a weakly supervised softmax cross-entropy loss is calculated on the labeled points as:
\begin{equation}
\label{segloss}
    \mathcal{L}_{seg\_basic} = -\frac{1}{\bm{B}}\displaystyle\sum_n \bm{m}_n\displaystyle\sum_{c} \bm{y}_{nc}\frac{e^{\bm{z}_{nc}}}{\displaystyle\sum_{c}e^{\bm{z}_{nc}}}
\end{equation}
where $\bm{B}=\displaystyle\sum_n \bm{m}_n$ is a normalization factor.

%-------------------------------------------------------------------------
\begin{figure}[t]
\begin{center}
   \includegraphics[width=0.9\linewidth]{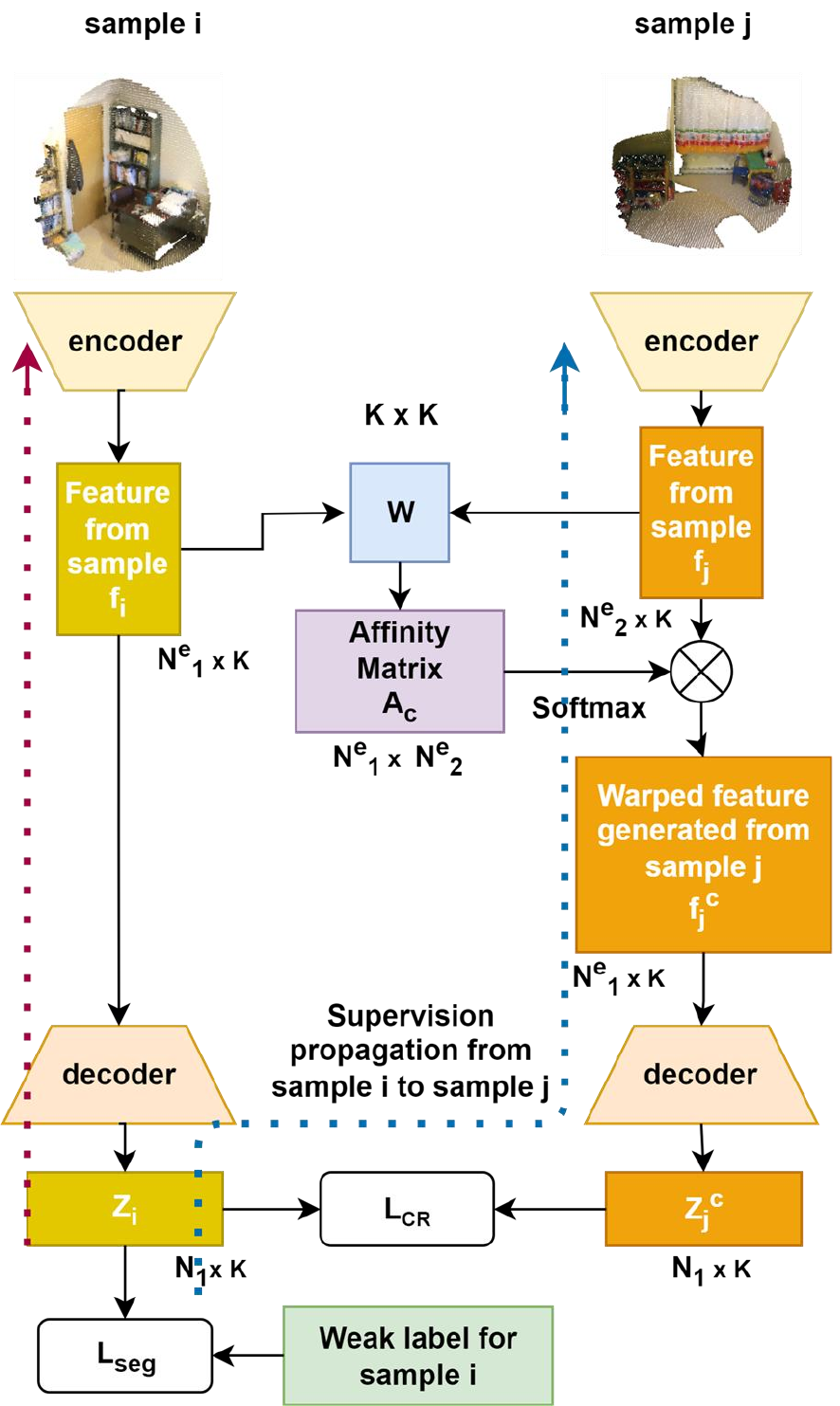}
\end{center}
   \caption{This blue dotted line illustrates how the weak supervision from sample $i$ is propagated to sample $j$ based on point correlation. For clarity, this figure shows only a single direction of the feature reallocating and supervision propagation, the module can propagate supervision bidirectionally.}
\label{fig:detail}
\vspace{-10pt}
\end{figure}

%-------------------------------------------------------------------------
\subsection{Cross-sample feature reallocating}\label{CSFR}
We take two point cloud samples $\bm{P}_i \in \mathbb{R}^{N_i \times D}$ and $\bm{P}_j \in \mathbb{R}^{N_j \times D}$ with at least one common label as an input pair. We feed the input pair into the KPConv encoder in the basic network and get the feature maps $\bm{F}_i \in \mathbb{R}^{N^e_i \times K}$ and $\bm{F}_j \in \mathbb{R}^{N^e_j \times K}$.

The feature reallocating is achieved by warping point features using the point correlation between the input samples. We first calculate a cross affinity matrix $\bm{A}_c$ between the two features. With a learnable matrix $\bm{W}_c \in \mathbb{R}^{K \times K}$, the affinity matrix is derived as:
\begin{equation}
    \bm{A}_c = \bm{F}_i \bm{W}_c \bm{F}^{\top}_j \in \mathbb{R}^{N^e_i \times N^e_j}
\end{equation}
The cross-affinity matrix describes the point-wise similarity between all pairs of points in two samples. We then normalize the cross affinity matrix row-wise and column-wise to guide the feature reallocating for each point:
\begin{equation}
\begin{split}
    &\bm{A}^i_c = softmax(\bm{A}_c)\\
    &\bm{A}^j_c = softmax(\bm{A}^{\top}_c)
\end{split}
\end{equation}
Then, we use the normalized affinity to guide the feature reallocating:
\begin{equation}
\begin{split}
    &\bm{F}^c_j = \bm{A}^i_c\bm{F_j} \in \mathbb{R}^{N^e_i \times K}\\
    &\bm{F}^c_i = \bm{A}^j_c\bm{F_i} \in \mathbb{R}^{N^e_j \times K}
\end{split}
\end{equation}
The warped feature maps $\bm{F}^c_j$ and $\bm{F}^c_i$ have the same size and spatial shape as the original feature maps $\bm{F}_i$ and $\bm{F}_j$. However, all the features in $\bm{F}^c_j$ are collected from sample $\bm{P}_j$ and each point feature in $\bm{F}^c_j$ can be seen as a weighted sum of all point features from $\bm{F}_j$, which means we constructed a new feature  $\bm{F}^c_j$ in the shape of $\bm{F}_i$ using the features from $\bm{F}_j$ and vice versa.

Unlike the previous methods\cite{sun2020mining, liu2020weakly, fan2020cian} which utilize common semantics from other samples and directly discover more object regions by generating pseudo labels. We feed the reallocated and original feature maps into the decoders with shared weights respectively. For sample $\bm{P}_i$ side, we feed the original feature map $\bm{F}_i$ and the reallocated feature map $\bm{F}^c_j$ from sample $\bm{P}_j$ into the decoder network and get the outputs $\bm{Z}_i = f_{dec}(\bm{F}_i) \in \mathbb{R}^{N_i \times C}$ and  $\bm{Z}^c_j = f_{dec}(\bm{F}^c_j) \in \mathbb{R}^{N_i \times C}$.

In our approach, the weak labels are assigned to a sample $\bm{P}_i$, where each point in the set $\bm{F}^c_j$ can be interpreted as a weighted sum derived from all points within sample $\bm{P}_i$. This characteristic allows for the dense propagation of supervision from labeled points in $\bm{P}_i$ to unlabeled points in 
$\bm{P}_j$, facilitated through the re-routing of gradients using our feature reallocating module. However, it is essential to consider the presence of uncommon classes across input samples. Directly applying a weak segmentation loss to the output of the reallocated branch might introduce undesirable noise during the training process.

%\textcolor{blue}{
To mitigate this challenge, we opt not to compute the segmentation loss for the reallocated branch. Instead, drawing inspiration from previous works such as \cite{tarvainen2017mean,chen2020simple,chen2020big}, we initially calculate a weak segmentation loss, as denoted in Equation \ref{segloss}, on the original feature output $\bm{Z}_i$. Additionally, we introduce a novel cross regularization (CR) loss designed to maximize the alignment between $\bm{Z}_i$ and $\bm{Z}^c_j$. This strategic combination of weak segmentation loss and cross regularization ensures a robust training framework, effectively addressing the challenge posed by the presence of uncommon classes in the input samples.
%}

\begin{equation}
    \mathcal{L}_{CR} = \|\bm{Z}_i - \bm{Z}^c_j\|^2_F = \|f_{dec}(\bm{F}_i)-f_{dec}(\bm{F}^c_j)\|^2_F
\end{equation}

Indeed, our methodology allows for the implementation of soft propagation, enabling the seamless transfer of supervision signals from labeled points in sample $\bm{P}_i$  to unlabeled points in sample $\bm{P}_j$ and vice versa. Consequently, each labeled point effectively disseminates its supervision signals to points within the other sample that exhibit similar features. This intricate mechanism ensures a comprehensive and nuanced understanding of the dataset, enhancing the ability of the model to learn intricate patterns and relationships within and between samples.
%}

%-------------------------------------------------------------------------
\subsection{Intra-sample feature reallocating}\label{ISFR}
In this section, we propose an intra-sample feature propagation(ISFR) module to further propagate supervision from labeled points to unlabeled points within each sample and finetuning the network against the possible noise introduced by the CSFR module. 
For each input sample $\bm{P}_i \in \mathbb{R}^{N_i \times D}$, we can get the feature map $\bm{F}_i \in \mathbb{R}^{N^e_i \times K}$ from the same encoder network. Similar to the previous module, we can calculate a self-affinity matrix $\bm{A}_s$ for the feature:
\begin{equation}
    \bm{A}_s = \bm{F}_i \bm{W}_s \bm{F}^{\top}_i \in \mathbb{R}^{N^e_i \times N^e_i}
\end{equation}
where $\bm{W}_s \in \mathbb{R}^{K \times K}$ is a learnable matrix.
The self-affinity matrix describes the point-wise correlation between all points within the input sample. We then normalize the self-affinity matrix to guide the feature reallocating for each point:
\begin{equation}
    \bm{A}^i_s = softmax(\bm{A}^{\top}_s)
\end{equation}
Then, we use the normalized affinity to guide the feature warping:
\begin{equation}
    \bm{F}^s_i = \bm{A}^i_s\bm{F_i} \in \mathbb{R}^{N^e_i \times K}\\
\end{equation}
Unlike the classical self-attention\cite{wang2018non}, we remove the residual connection to retain the same activation intensity. Here each point in $\bm{F}^s_i$ can be considered as a weighted sum of all the points in the sample and the features are reallocated with respect to the point correlation. We separately feed the reallocated feature and the original feature to the decoder. Thus, the decoder outputs from the two branches are  $\bm{Z}^s_i = f_{dec}(\bm{F}^s_i) \in \mathbb{R}^{N_i \times C}$ and $\bm{Z}_i = f_{dec}(\bm{F}_i) \in \mathbb{R}^{N_i \times C}$.

Similar to the cross-regularization loss, we use a similar self-regularization (SR) loss on the two outputs to regularize the reallocated feature with the original output:
\begin{equation}
    \mathcal{L}_{SR} = \|\bm{Z}^s_i - \bm{Z}_i\|^2_F = \|f_{dec}(\bm{F}^s_i)-f_{dec}(\bm{F}_i)\|^2_F.
\end{equation}

Given that our feature reallocation process occurs exclusively within samples and does not introduce supervision from absent classes, unlike methods such as CSFR, we employ an additional weak segmentation loss specifically on the self-reallocating branch.
\begin{equation}
    \mathcal{L}_{seg\_s} = -\frac{1}{\bm{B}}\displaystyle\sum_n \bm{m}_n\displaystyle\sum_{c} \bm{y}_{nc}\frac{e^{\bm{z}^s_{nc}}}{\displaystyle\sum_{c}e^{\bm{z}^s_{nc}}}.
\end{equation}

here $\bm{B}=\displaystyle\sum_n \bm{m}_n$ represents a normalization factor. This normalization factor plays a crucial role, enabling the effective transfer of supervision signals from labeled points to unlabeled points with similar features.

%-------------------------------------------------------------------------
\subsection{Training}\label{training}
As shown in Figure\ref{fig:framework}, we use a two-stage training strategy to avoid interference between the two modules during training. In stage one, we train the basic segmentation network with the cross-sample feature reallocating module. In this stage, for each sample, the network learns from the weak labels of this sample and the supervision propagated from the weak labels of another sample.
 
The overall learning objective in stage one can be expressed as:
\begin{equation}
    \mathcal{L}_{stage1} = \mathcal{L}_{seg\_basic} + \mathcal{L}_{CR}
\end{equation}
In the second stage, we train the basic segmentation network with the intra-sample feature reallocating module. The second stage propagates supervision from labeled points to unlabeled points. The overall training objectives in stage two are:
\begin{equation}
    \mathcal{L}_{stage2} = \mathcal{L}_{seg\_basic} +  \mathcal{L}_{seg\_s} + \mathcal{L}_{SR}
\end{equation}
We argue that joint training of the two modules would hamper the performance of the basic segmentation network since the cross-regularization loss in the CSFR module and the self-regularization loss in the ISFR module may interfere with the training of each other as the CSFR module may bring wrong supervision from uncommon classes to the other sample, while the two-stage training can avoid this issue and improve the performance by taking advantage of both the modules. We will further discuss this through experiments in section \ref{ablation}. 

The two modules implicitly guide the optimization of the basic network only at training time. In inference, we can simply discard the two modules and use the basic segmentation network as a normal point cloud segmentation network to get the segmentation predictions. Therefore, no extra memory and computational resources are introduced at test time.

%-------------------------------------------------------------------------
\begin{table}[b]
\caption{Ablation study on one stage and two stage training. We evaluate each module in one-stage training and the training orders in two-stage training. Evaluated on S3DIS Area-5 in mIoU(\%).}
%\begin{center}
\centering
\setlength\extrarowheight{3pt}
\resizebox{0.7\linewidth}{!}{\begin{tabular}{lcc}
\hline
\multicolumn{1}{l|}{modules}   & \multicolumn{1}{c|}{10\% label} & 1\% label \\ \hline
\multicolumn{1}{l|}{Baseline}  & \multicolumn{1}{c|}{66.5}       & 65.1      \\ \hline
\multicolumn{3}{c}{One stage training}                                       \\ \hline
\multicolumn{1}{l|}{CSFR}      & \multicolumn{1}{c|}{67.7}       & 66.8      \\
\multicolumn{1}{l|}{ISFR}      & \multicolumn{1}{c|}{68.0}       & 66.9      \\
\multicolumn{1}{l|}{CSFR+ISFR} & \multicolumn{1}{c|}{66.3}       & 65.0      \\ \hline
\multicolumn{3}{c}{Two stage training}                                       \\ \hline
\multicolumn{1}{l|}{ISFR-CSFR} & \multicolumn{1}{c|}{67.3}       & 65.4      \\
\multicolumn{1}{l|}{CSFR-ISFR} & \multicolumn{1}{c|}{\textbf{68.6}}       & \textbf{67.0}      \\ \hline
\end{tabular}}
%\end{center}

\label{Tab:twostage}
\vspace{-10pt}
\vspace{-5pt}
\end{table}

%------------------------------------------------------------------------
\section{Experiments}

%-------------------------------------------------------------------------
\subsection{Dataset}
\textbf{Dataset:} We conduct our experiments on the popular public dataset S3DIS\cite{Armeni_2016_CVPR}. S3DIS covers 6 areas of the entire floor from 3 different buildings with a total of 215 million points and covers over 6000$m^2$. The dataset is annotated with 13 classes. We follow the common practice that using Area 5 as the test scene to measure the generalization ability. We also perform experiments on ScanNet\cite{dai2017scannet} which contains 1513 scenes and is annotated with 20 classes.

\textbf{Weak labels: } We follow \cite{xu2020weakly} to annotate only 10\% of the points. We first sample 4\% of the points from the original data as the network inputs. Then, we randomly label 10\% of points in each class for the sampled input point clouds. The final predictions will be back-projected to the original point clouds. Therefore, only 10\% of the network input training data and only 0.4\% of the original point cloud data are labeled. We also perform experiments with fewer labels that only 1\% of the input points are labeled.

%-------------------------------------------------------------------------
\subsection{Implementation details}
We use the KPConv\cite{thomas2019kpconv} segmentation model KPFCNN as our backbone network. The network is an encoder-decoder fully convolutional network with skip connections. The encoder is composed by bottleneck ResNet blocks\cite{he2015deep} with KP convolution layers. The decoder part is composed of the nearest upsampling layers with unary convolution layers. We put the CSFR and ISFR modules after the first upsampling layer for larger spatial resolution. Due to the limitation in computational resources, we use ball query to sample point cloud as input samples, the sample radius is set to 2m. We use a Momentum SGD optimizer, the initial learning rate is set to 0.01 and the momentum is set to 0.98. We train the first stage for 600 epochs and the second stage for another 600 epochs. 

In this study, the comprehensive duration of the training process, encompassing both stages, is approximately 33 GPU hours when executed on a single GTX2080Ti GPU. This duration, while marginally lengthier than the training time reported in the WeakSup\cite{xu2020weakly}, is notably shorter than the training time in the One-Thing-One-Click\cite{liu2021one}. This brevity arises from the utilization of simpler backbone networks in WeakSup and the necessity for iterative training on the same network in One-Thing-One-Click. Notably, the inference duration of our proposed method aligns precisely with that of the fully supervised KPConv\cite{thomas2019kpconv}. This parity in inference times stems from the exclusive involvement of the CSFR and ISFR modules solely during the training phase, while the network architecture during inference mirrors that of the KPConv model.

\begin{table}[]
\centering
\caption{Training time in GPU hours on a Single GTX2080Ti.}
\label{tab:gouhour}

\resizebox{0.9\columnwidth}{!}{%
% \color{blue}
\begin{tabular}{c|ccc}

\hline

Method  & Dataset & Backbone   & GPU Hour \\ \hline
WeakSup\cite{xu2020weakly} & S3DIS   & DGCNN\cite{wang2019dynamic}      & 26.65    \\
Ours     & S3DIS   & KPConv\cite{thomas2019kpconv}     & 33.85    \\
OTOC\cite{liu2021one}    & ScanNet & PointGroup\cite{jiang2020pointgroup} & 175      \\
Ours     & ScanNet & KPConv\cite{thomas2019kpconv}     & 33.32    \\ \hline

\end{tabular}%
}
\end{table}
%change
%-------------------------------------------------------------------------
\begin{table}[t]
\caption{Ablation study on different losses within each module in mIoU(\%). (a) illustrates the effectiveness of losses in the CSFR module. We conduct experiments with an additional $\mathcal{L}_{seg\_c}$ which is not included in the proposed method for comparison. (b) illustrates the effectiveness of losses in the ISFR module.}
\centering
\setlength\extrarowheight{3pt}
\resizebox{0.85\linewidth}{!}{\begin{tabular}{ccccc}
\hline
\multicolumn{5}{c}{CSFR}                  \\ \hline
$\mathcal{L}_{seg\_basic}$ & $\mathcal{L}_{CR}$ & $\mathcal{L}_{seg\_c}$ & 10\%          & 1\%           \\ \hline
\checkmark &   &   & 66.5          & 65.1          \\
\checkmark & \checkmark &   & \textbf{67.7} & \textbf{66.8} \\
\checkmark & \checkmark & \checkmark & 66.3          & 64.8          \\ \hline 
\multicolumn{5}{c}{(a)} \\ \hline
\multicolumn{5}{c}{ISFR}                  \\ \hline
$\mathcal{L}_{seg\_basic}$ & $\mathcal{L}_{SR}$ & $\mathcal{L}_{seg\_s}$ & 10\%          & 1\%           \\ \hline
\checkmark &   &   & 66.5          & 65.1          \\
\checkmark & \checkmark &   & 66.9          & 65.9          \\
\checkmark & \checkmark & \checkmark & \textbf{68.0} & \textbf{66.9} \\ \hline
\multicolumn{5}{c}{(b)}
\end{tabular}}

\label{tab: loss}
\vspace{-10pt}
\vspace{-5pt}
\end{table}

%-------------------------------------------------------------------------
\begin{table}[]
\caption{We compare the segmentation predictions in mIoU(\%) from different decoder branches for the two stages of training. The basic branch is the one we use for inference. Cross-branch and intra-branch are the decoder outputs produced from CSFR and ISFR propagated features.}
\centering
\setlength\extrarowheight{5pt}
\resizebox{0.99\linewidth}{!}{\begin{tabular}{c|ccc}
\hline
setting   & basic branch & cross branch & intra branch \\ \hline
ISFR-CSFR & 67.3         & 60.3        & 66.6        \\ \hline
CSFR-ISFR & \textbf{68.6}         & 62.2        & 67.4        \\ \hline
\end{tabular}}

\label{tab:branches}
\vspace{-10pt}
\vspace{-5pt}
\end{table}
%-------------------------------------------------------------------------
\subsection{Ablation studies}\label{ablation}

%-------------------------------------------------------------------------
\textbf{Two stage training versus joint training: }Table \ref{Tab:twostage} compares one-stage training with two-stage training performances trained with 10\% and 1\% labels. For one-stage training, we perform experiments with only the CSFR module or ISFR module, each of the modules produces performance gain over the baseline method for both 10\% and 1\% label cases. However, when we jointly train CSFR and ISFR modules in one stage, we observe a performance drop, producing results lower than solely training one module and even lower than the baseline method in both cases. From the experiments, we argue that the training losses in the two modules may interfere with each other during optimization. Since the CR loss and SR loss from the two modules both trying to pull the segmentation output closer to the outputs of the two branches, the overall training objective may be deviated from our real target, better segmentation performance. With the losses from two modules being optimized together, the model may fail to find the global optimum and be optimized in the wrong direction. Therefore, by separating the training into two stages, we can avoid interference between the modules during training and take advantage of both modules.
%-------------------------------------------------------------------------
\begin{figure*}[t]
\begin{center}
   \includegraphics[width=0.99\linewidth]{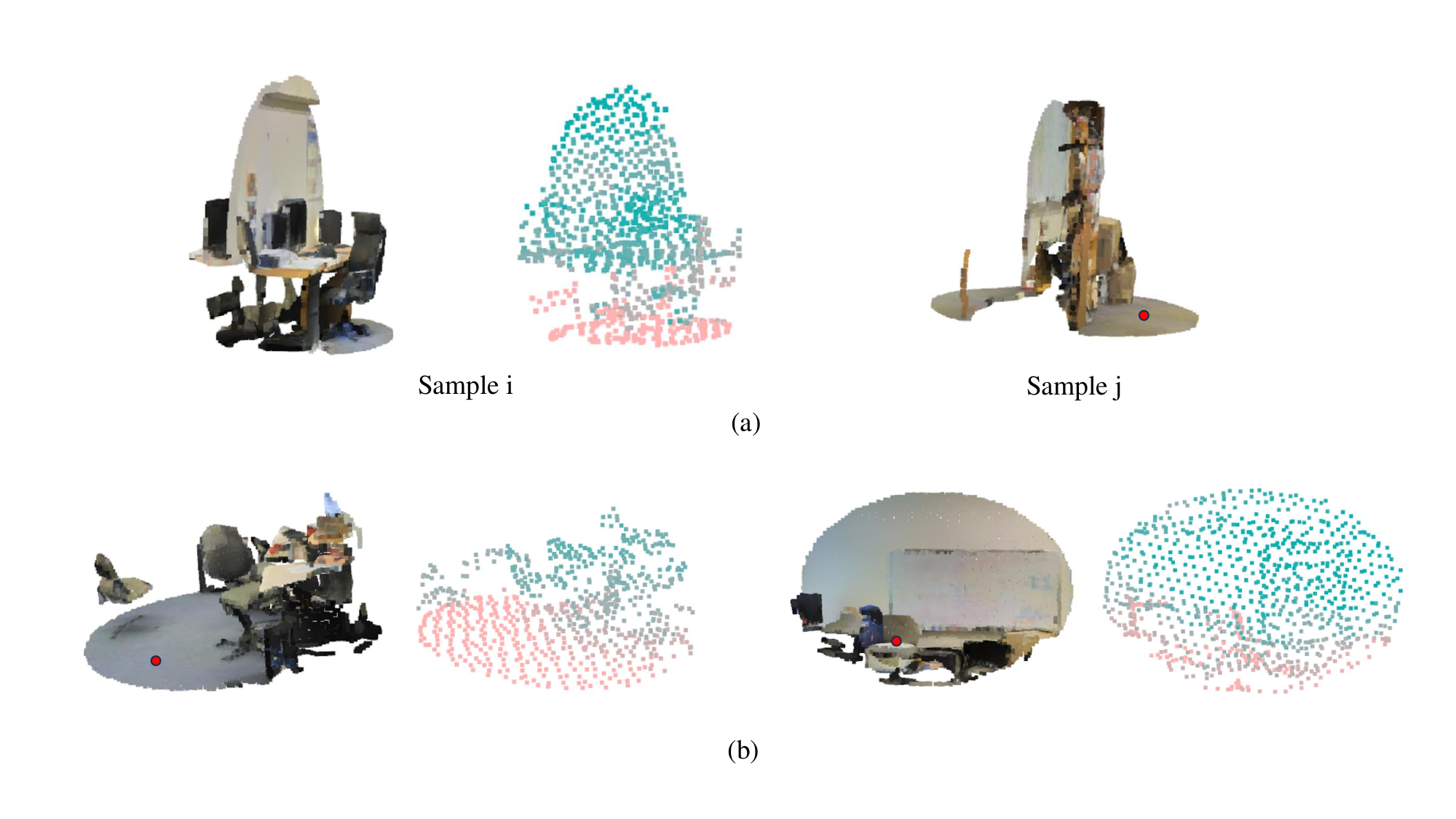}
\end{center}
   \caption{The visualization for the affinity calculated by (a) CSFR module, (b) ISFR module. The colors on the point cloud show the corresponding affinity of each point to the point indicated by the red dot. Pink means higher similarity and green means lower similarity.}
\label{fig:crossaff}
%\vspace{-20pt}
\vspace{-15pt}
\end{figure*}

%-------------------------------------------------------------------------
\begin{figure*}[t]
\begin{center}
%\fbox{\rule{0pt}{2in} \rule{.9\linewidth}{0pt}}
\includegraphics[width=0.99\linewidth]{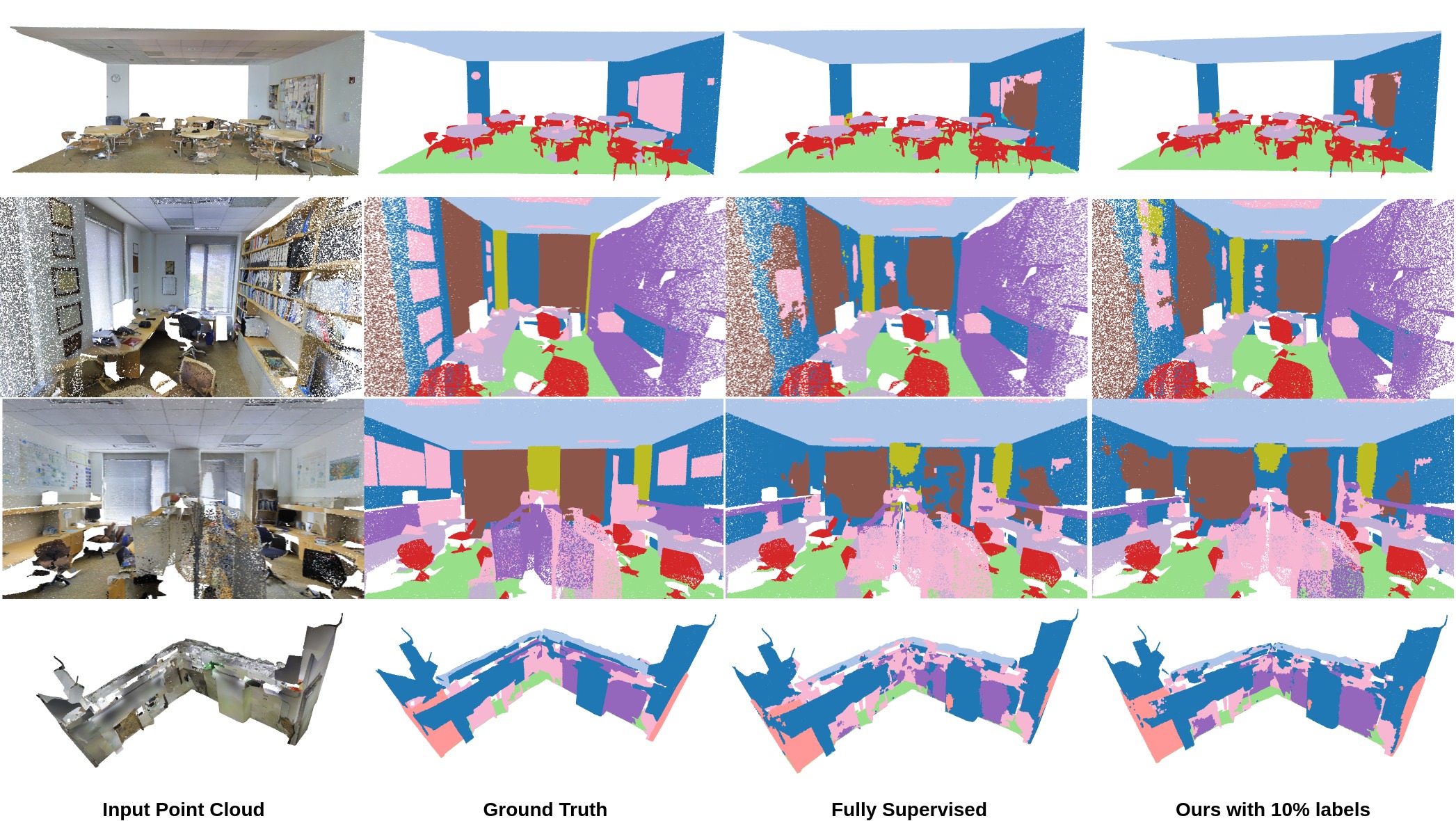}
\end{center}
   \caption{Visualizations on S3DIS Area-5. The results are, left to right, input RGB point cloud, ground truth, fully supervised method, and our proposed method with 10\% labels.}
\label{fig:visfinal}
%\vspace{-15pt}
\vspace{-15pt}
\end{figure*}
%-------------------------------------------------------------------------
\textbf{Training orders: }We also compare the training orders of the two modules in Table \ref{Tab:twostage}. ISFR-CSFR means we train the network with the ISFR module in the first stage and the CSFR module in the second stage. CSFR-ISFR means that we train the network with the CSFR module in the first stage and ISFR module in the second stage. We observe that CSFR-ISFR outperforms ISFR-CSFR under both 10\% and 1\% supervision. We argue that the CSFR module can help the network to learn more generalized coarse representations across samples while the ISFR module can act like a finetuning module that finetunes the learned representations and imposes constraints on unlabeled points within the samples. Therefore if we use the opposite training order, we can get worse results compared to even each single module.

%-------------------------------------------------------------------------
\textbf{Effects of different losses: }We evaluate the effectiveness of each losses in Table \ref{tab: loss}
within each module. We compare results with different combinations of the losses for a single-stage training with only one module each. For the CSFR module, as shown in Table \ref{tab: loss}(a), we suppose $\mathcal{L}_{seg\_c}$ is a segmentation loss appended to the decoder of the cross-propagated branch which we did not include in our architecture. We can observe that training with $\mathcal{L}_{seg\_basic} + \mathcal{L}_{CR}$ reaches the best performance for the CSFR module for both the label settings which means the cross regularization loss helps the segmentation performance. We also observe that training with $\mathcal{L}_{seg\_basic} + \mathcal{L}_{CR} + \mathcal{L}_{seg\_c}$ produces even lower score than solely using $\mathcal{L}_{seg\_basic}$, which means adding another segmentation loss on the cross propagated decoder branch would harm the performance. This supports our statement in section \ref{CSFR} that additional supervision on the propagate branch would introduce noise into the training process due to the inherent difference between the two samples.

Table \ref{tab: loss}(b) shows that in the ISFR branch, training with all three losses $\mathcal{L}_{seg\_basic} + \mathcal{L}_{SR} + \mathcal{L}_{seg\_s}$ produces better results than using $\mathcal{L}_{seg\_basic} + \mathcal{L}_{SR}$ and solely using the segmentation loss on the original branch. This result supports our statement in section \ref{ISFR} that the ISFR module imposes constraints on unlabeled points within the sample. Since the feature propagation is processed inside each sample, no noises would be introduced to the training. Thus, calculating a segmentation loss on the propagated feature branch would improve the training results.

%-------------------------------------------------------------------------
\begin{table*}[ht]
\caption{The class-specific mIoU (\%) evaluation on S3DIS Area-5. KPConv(paper) is taken from the paper-reported score, and KPConv(retrain) is the score from our basic segmentation network trained with 100\% labels. The baseline method means the basic segmentation network trained with only the weak labels and without CSFR and ISFR modules.}
\centering
\setlength\extrarowheight{3pt}
%\color{blue}
\resizebox{0.99\linewidth}{!}{
\begin{tabular}{ccccccccccccccccc}
\hline

\multicolumn{1}{c|}{setting}        & \multicolumn{1}{c|}{model}      &   \multicolumn{1}{c|}{backbone} & \multicolumn{1}{c|}{mIoU} & ceil. & floor & wall & beam & col. & wind. & door & chair & table & book. & sofa & board & clut. \\ \hline
\multicolumn{16}{c}{Fully supervised}                                                                                                                                                                \\ \hline
\multicolumn{1}{c|}{}               & \multicolumn{1}{c|}{PointNet\cite{qi2017pointnet}}& \multicolumn{1}{c|}{-}   & \multicolumn{1}{c|}{41.1} & 88.8  & 97.3  & 69.8 & 0.1  & 3.9  & 46.3  & 10.8 & 52.6  & 58.9  & 40.4  & 5.9  & 26.4  & 33.2  \\
\multicolumn{1}{c|}{}               & \multicolumn{1}{c|}{PointNet++\cite{qi2017pointnet++}}& \multicolumn{1}{c|}{-} & \multicolumn{1}{c|}{47.8} & 90.3  & 95.6  & 69.3 & 0.1  & 13.8 & 26.7  & 44.1 & 64.3  & 70.0  & 27.8  & 47.8 & 30.8  & 38.1  \\
\multicolumn{1}{c|}{}               & \multicolumn{1}{c|}{DGCNN\cite{wang2019dynamic}}& \multicolumn{1}{c|}{-}      & \multicolumn{1}{c|}{47.0} & 92.4  & 97.6  & 74.5 & \textbf{0.5}  & 13.3 & 48.0  & 23.7 & 65.4  & 67.0  & 10.7  & 44.0 & 34.2  & 40.0  \\
\multicolumn{1}{c|}{}               & \multicolumn{1}{c|}{PointCNN\cite{li2018pointcnn}}& \multicolumn{1}{c|}{-}   & \multicolumn{1}{c|}{57.3} & 92.3  & 98.2  & 79.4 & 0.0  & 17.6 & 22.8  & 62.1 & 80.6  & 74.4  & 66.7  & 31.7 & 62.1  & 56.7  \\
\multicolumn{1}{c|}{}               & \multicolumn{1}{c|}{MinkNet\cite{choy20194d}}& \multicolumn{1}{c|}{-}    & \multicolumn{1}{c|}{65.4} & 91.8  & \textbf{98.7}  & \textbf{86.2} & 0.0  & \textbf{34.1} & 48.9  & 62.4 & 89.8  & 81.6  & 74.9  & 47.2 & 74.4  & 58.6  \\
\multicolumn{1}{c|}{}               & \multicolumn{1}{c|}{JSENet\cite{hu2020jsenet}} & \multicolumn{1}{c|}{-}    & \multicolumn{1}{c|}{67.7} & 93.8  & 97.0  & 81.4 & 0.0  & 23.2 & \textbf{61.3}  & 71.6 & 89.9  & 79.8  & \textbf{75.6}  & 72.3 & 72.7  & 60.4  \\
\multicolumn{1}{c|}{Paper}          & \multicolumn{1}{c|}{KPConv\cite{thomas2019kpconv}} & \multicolumn{1}{c|}{-}    & \multicolumn{1}{c|}{67.1} & 92.8  & 97.3  & 82.4 & 0.0  & 23.9 & 58.0  & 69.0 & 91.0  & 81.5  & 75.3  & \textbf{75.4} & 66.7  & 58.9  \\
\multicolumn{1}{c|}{Retrain}        & \multicolumn{1}{c|}{KPConv\cite{thomas2019kpconv}} & \multicolumn{1}{c|}{-}    & \multicolumn{1}{c|}{67.3} & \textbf{94.8}  & 98.4  & 82.3 & 0.0  & 28.3 & 56.6  & 71.2 & 90.5  & \textbf{82.6}  & 74.6  & 67.8 & 67.4  & 59.9  \\ \hline
\multicolumn{1}{c|}{Fully supervised}        & \multicolumn{1}{c|}{DSP(Ours)} & \multicolumn{1}{c|}{KPConv\cite{thomas2019kpconv}}   & \multicolumn{1}{c|}{\textbf{68.5}} & 94.3  & 98.4  & 83.7 & 0.0  & 28.9 & 59.1  & \textbf{73.4} & \textbf{91.2}  & 82.2  & 75.3  & 73.9 & \textbf{68.5}  & \textbf{61.0}  \\ \hline

\multicolumn{16}{c}{Weakly supervised}                                                                                                                                                               \\ \hline
\multicolumn{1}{c|}{2D dense label} & \multicolumn{1}{c|}{GPFN\cite{wang2020weakly}}   & \multicolumn{1}{c|}{-}      & \multicolumn{1}{c|}{50.8} & -     & -     & -    & -    & -    & -     & -    & -     & -     & -     & -    & -     & -     \\
%change

\multicolumn{1}{c|}{10\% label}     & \multicolumn{1}{c|}{$\Pi$ Model\cite{laine2016temporal}}& \multicolumn{1}{c|}{-}  & \multicolumn{1}{c|}{43.6} &  91.8 & 97.1 & 73.8 & 0.0 & 5.1 & 42.0 & 19.6 & 66.7 & 67.2 & 19.1 & 47.9 & 30.6 & 41.3  \\ 
\multicolumn{1}{c|}{10\% label}     & \multicolumn{1}{c|}{MT\cite{tarvainen2017mean}} & \multicolumn{1}{c|}{-} & \multicolumn{1}{c|}{47.9} & 92.2 & 96.8 & 74.1 & 0.0 & 10.4 & 46.2 &17.7 & 67.0 & 70.7 & 24.4 & 50.2 & 30.7 & 42.2  \\ 
\multicolumn{1}{c|}{10\% label}     & \multicolumn{1}{c|}{WeakSup\cite{xu2020weakly}}& \multicolumn{1}{c|}{PointNet\cite{qi2017pointnet}}  & \multicolumn{1}{c|}{48.0} & 90.9  & 97.3  & 74.8 & 0.0  & 8.4  & 49.3  & 27.3 & 69.0  & 71.7  & 16.5  & 53.2 & 23.3  & 42.8  \\ 
\multicolumn{1}{c|}{10\% label}     & \multicolumn{1}{c|}{OTOC\cite{liu2021one}} & \multicolumn{1}{c|}{MinkNet\cite{choy20194d}} & \multicolumn{1}{c|}{48.2} & 91.2 & 97.7 & 78.0 & 0.0 & 6.3 & 46.3 & 31.6 & 65.7 & 64.4 & 8.2 & 52.5 & 41.6 & 43.1  \\ 
\multicolumn{1}{c|}{10\% label}     & \multicolumn{1}{c|}{MulPro\cite{su2023weakly}} & \multicolumn{1}{c|}{DGCNN\cite{wang2019dynamic}} & \multicolumn{1}{c|}{49.0} & 89.7 & 96.9 & 75.5 & 0.0 & 14.0 & 45.7 & 40.7 & 68.5 & 66.8 & 13.9 & 49.4 & 34.4 & 41.2  \\ 
%change
\hline
\multicolumn{1}{c|}{1\% label}      & \multicolumn{1}{c|}{Baseline}   & \multicolumn{1}{c|}{KPConv\cite{thomas2019kpconv}} & \multicolumn{1}{c|}{65.1} & 93.8  & 97.9  & 81.8 & 0.0  & 28.6 & 57.1  & 67.6 & 89.1  & 79.7  & 73.4  & 59.7 & 65.9  & 59.9  \\
\multicolumn{1}{c|}{1\% label}      & \multicolumn{1}{c|}{DSP(Ours)}   & \multicolumn{1}{c|}{KPConv\cite{thomas2019kpconv}}     & \multicolumn{1}{c|}{67.0} & 93.5  & 98.2  & 82.8 & 0.0  & 29.3 & \textbf{58.3}  & 68.7 & 90.5  & 80.0  & 72.2  & 74.1 & 65.4  & 58.0  \\
\multicolumn{1}{c|}{10\% label}     & \multicolumn{1}{c|}{Baseline}   & \multicolumn{1}{c|}{KPConv\cite{thomas2019kpconv}} & \multicolumn{1}{c|}{66.5} & \textbf{94.5}  & \textbf{98.4}  & 82.9 & 0.0  & 26.8 & 56.3  & 71.8 & 90.6  & 82.0  & \textbf{75.4}  & 59.7 & 65.9  & 59.9  \\
\multicolumn{1}{c|}{10\% label}     & \multicolumn{1}{c|}{DSP(Ours)}   & \multicolumn{1}{c|}{KPConv\cite{thomas2019kpconv}}     & \multicolumn{1}{c|}{\textbf{68.6}} & 93.6  & 98.3  & \textbf{83.6} & 0.0  & \textbf{31.1} & 57.8  & \textbf{72.1} & \textbf{91.1}  & \textbf{82.1}  & 75.2  & \textbf{75.9} & \textbf{70.3}  & \textbf{60.5}  \\ \hline

\end{tabular}
}
\vspace{-10pt}

\label{tab:Results}
%\vspace{-15pt}
\end{table*}
%-------------------------------------------------------------------------
%-------------------------------------------------------------------------
\textbf{Performance of different branches: }Table \ref{tab:branches} compare the segmentation performance of different decoder branches in the two-stage settings. In both settings, the cross branches produce the poorest segmentation results the features of this branch are propagated from the other sample. However, the cross-branch can still produce reasonable scores as the network is learned to only propagate features from the same category for each point. The intra branch produces better results than the cross branch, but still lower than the basic branch. This supports our argument that the ISFR module imposes more constraints to unlabeled points as the features of unlabeled points can be propagated to the labeled points in the same category. But during inference, as the network already learned the representations, propagating the features is not helping the predictions. This is also a difference between our ISFR module with the self-attention module in many vision applications.

We also compare the results between the two settings. The outputs from all three branches in CSFR-ISFR perform better than those in ISFR-CSFR, which also supports our argument in section \ref{training}. From the results, if the ISFR module is trained first, the network can overfit within samples and not generalize across samples. Thus the cross-branch performance is worse in ISFR-CSFR. Then, with the CSFR module trained in the second stage, the information from other samples may affect as noises. Therefore the performance from the basic branch is even lower than individual training with the two modules shown in Table \ref{Tab:twostage}.
%-------------------------------------------------------------------------

\begin{table*}[]
        \centering
        \setlength{\abovedisplayskip}{0pt}
        \setlength{\belowdisplayskip}{0pt}
        %\small
        %\tiny
        %\scriptsize
        \caption{The segmentation results on ScanNet validation set. MPRM is another weakly supervised method using sub-cloud level supervision and developed also based on KPConv. The KPConv-basic means the basic network developed using KPConv.}
        %\footnotesize
        \setlength\tabcolsep{1pt}
        \setlength\extrarowheight{3pt}

        \resizebox{0.99\linewidth}{!}{\begin{tabular}{l|c|cccccccccccccccccccc|c}
            \hline

             Model & Supervision &wall &floor &cabinet & bed & chair &sofa & table &door &window &B.S. & picture & counter &desk &curtain & fridge & S.C. &toilet & sink & bathtub & other & mIoU \\
            \hline \hline

            MPRM\cite{wei2020multi} & subcloud &  59.4& 59.6& 25.1& 64.1& 55.7& 58.7& 45.6& 36.4& 40.3& 67.0& 16.1& 22.6& 42.9& 66.9& 24.1& 39.6& 47.0& 21.2& 44.7& 28.0& 43.2 \\%[1.5pt]
            \hline

             KPConv-basic  & Full&  83.9  & 95.5  & \textbf{68.5}  &79.4  &\textbf{90.0} &  80.9&      \textbf{77.4} &  63.0 &  \textbf{60.8}  & 81.0 & 29.6 &  \textbf{68.3} &  \textbf{67.6} &  \textbf{77.3} &     54.5 & 61.1 & 90.9  &  62.9 &  86.4 &  56.8 & 71.8 \\%[1.5pt]
             KPConv-basic  & 1\% & 82.7&  94.9 & 62.8 & 77.8 & 87.7& 78.8 & 74.0 & 58.7 & 55.8 & 75.9 & 30.5 & 55.2 & 63.2 & 69.2 & 48.6 & 62.6 & 87.3 & 57.8 & 84.4 & 54.4 & 68.1 \\%[1.5pt]
             KPConv-basic & 10\%  &  83.6 & 95.1 & 66.1 & 79.9 & 89.3 & 82.5 & 74.1 & 62.1 & 60.5 & 80.1 & \textbf{32.5} & 61.2 & 64.1 & 73.3 & 55.7 & 63.3 & 89.7 & \textbf{63.4} & \textbf{89.7} & \textbf{63.4}& 71.1 \\%[1.5pt]
             DSP (Ours) & 1\% & 82.8  & 95.4  & 64.7 & 78.8 & 89.1 & \textbf{82.5} &      76.2 & 61.1 & 57.0  & 79.1 &  30.8 &  63.8 & 64.9  & 73.8  &   52.7   &   \textbf{66.8}   &  88.1   &  59.1 &  80.1  &  53.0   & 70.0\\%[1.5pt] 
             
             DSP (Ours) & 10\% & \textbf{84.2} & \textbf{95.5}  & 68.1 & \textbf{80.0} & 89.9 & 81.1 &    76.7  &  \textbf{64.2} & 60.7  & \textbf{81.3} & 31.3  & 68.1  & 66.0  & 75.9 &     \textbf{60.6} & 61.4 & \textbf{90.9} & 63.3 &  86.4 &  56.8   & \textbf{72.1}\\
             \hline

        \end{tabular}}

        \label{tab:scannet}
\vspace{-15pt}
    \end{table*}

%------------------------------------------------------------------------\

%-------------------------------------------------------------------------
\subsection{Qualitative results}
Figure\ref{fig:crossaff}(a) shows the affinity learned in CSFR between the input pairs. Figure\ref{fig:crossaff}(b) shows the affinity learned in ISFR in each sample. We show the affinity map on the left to the selected point on the right. The point clouds are sparse since we perform CSFR and ISFR in down-sampled features. Obviously, both modules have successfully learned to propagate features from the same class.

We provide visualizations of our final segmentation predictions in Figure\ref{fig:visfinal}. We compare our results with the ground truth and the predictions from the fully supervised counterpart. We can see that our method performs even better in label consistency and in hard examples like the fourth row.
%-------------------------------------------------------------------------

\subsection{Segmentation results}
In Table \ref{tab:Results}, we compare our proposed method with state-of-the-art methods, other weakly supervised methods, and our baseline method.

\textbf{Comparison with baseline method: }The baseline method is only the basic segmentation network trained with weak labels. Our method is the full model with cross- and intra-sample feature propagation modules and is trained with the two-stage strategy. 

The experiments show that our proposed method improves the performance of the baseline method with both 10\% and 1\% labels.

\textbf{Comparison with existing 3D WSSS methods: } 
% \textcolor{blue}{
We compare our proposed method with existing 3D WSSS methods\cite{xu2020weakly, laine2016temporal, tarvainen2017mean}. For One-Thing-One-Click\cite{liu2021one} and MulPro\cite{su2023weakly}, we use the results reported from the MulPro\cite{su2023weakly} paper. \cite{wang2020weakly} utilizes 2D dense labels on 2D projections of the 3D point clouds and \cite{xu2020weakly} utilize the same weak supervision technique by annotating 10\% of the points. \cite{xu2020weakly} can also produce close or better results than its fully supervised counterpart, but the overall performance of these existing methods\cite{xu2020weakly, laine2016temporal, tarvainen2017mean, su2023weakly} still remains a huge gap with existing state-of-the-art methods\cite{thomas2019kpconv, choy20194d}.
 Our proposed method surpasses the existing methods\cite{su2023weakly, su2023weakly, xu2020weakly} by a large margin.
% }

\textbf{Comparison with fully supervised methods: }We compare our weakly supervised method with some fully supervised state-of-the-art methods\cite{qi2017pointnet,qi2017pointnet++,wang2019dynamic,li2018pointcnn, choy20194d, li2018pointcnn, thomas2019kpconv} on the public dataset S3DIS Area-5. Also, we compare our method under weak supervision against full supervision. Our method produces even slightly higher results using 10\% label than trained under full supervision by 0.1\%. We believe that supervision propagation is not that useful under full supervision as each point can directly get supervision information from its own label but it can still benefit from the self-attention like architecture. Therefore our 10\% label results can produce similar results to the fully supervised counterpart.

\textbf{Results on ScanNet: }Table \ref{tab:scannet} shows the class specific segmentation results in mIoU(\%) in ScanNet\cite{dai2017scannet} validation set. Our method outperforms the previous method MPRM\cite{wei2020multi} by a large margin. We observe that with 10\% points labeled, the baseline results are already very close to the fully supervised results with only 0.7\% margin. We argue that ScanNet is a very large-scale dataset, 10\% of the label can already provide strong supervision. Our method with 10\% points labeled improved 1\% mIoU from the baseline and even outperformed the fully supervised result by a small margin of 0.3\%. Since the 10\% baseline is already close to the fully supervised result, our supervision propagation mechanism may help the network with more information than the fully supervised baseline model. Our method is more effective with only 1\% labeled points, where we outperform the baseline method by 1.9\% mIoU. In practice, it is not easy to collect a large-scale dataset like ScanNet, but we believe our method is meaningful in real-world applications. 

%-------------------------------------------------------------------------
\begin{figure}[]
\begin{center}
%\fbox{\rule{0pt}{2in} \rule{.9\linewidth}{0pt}}
\includegraphics[width=0.99\linewidth]{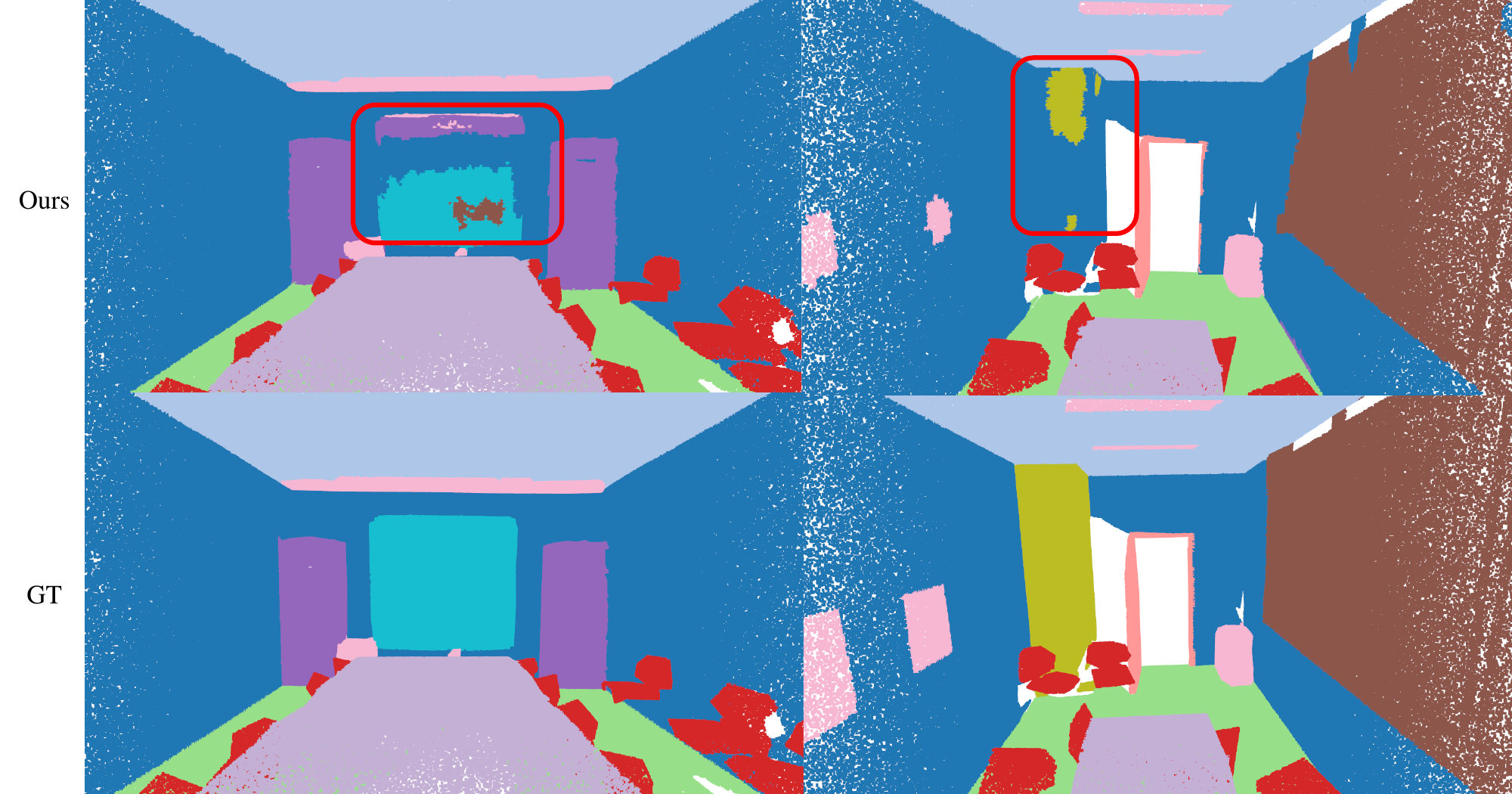}
\end{center}
   \caption{
   %minor
   Failure cases on S3DIS Area-5. We observed that our model encounters difficulties in distinguishing between categories that have geometric similarities. Specifically, when two categories exhibit close geometric resemblances, the model often struggles to delineate clear boundaries between them.}
\label{fig:failure}
%\vspace{-15pt}
\vspace{-15pt}
\end{figure}

%minor
%\textcolor{blue}{
\section{Limitations}
We present several failure cases in Figure \ref{fig:failure}. Consistent with numerous existing studies, our approach struggles when two categories exhibit similar geometric properties. This specific problem is a well-documented challenge, and its presence is notable even in methods that employ full supervision. The inherent limitations of our method, stemming from its lack of comprehensive supervision, further exacerbate its ability to effectively navigate and resolve this particular challenge. It underscores the need for enhanced techniques or additional guidance to improve differentiation in such cases.%}

\vspace{-10pt}

\section{Conclusion}
%\vspace{-5pt}
In this paper, we propose a weakly supervised point cloud segmentation method with only 10\% or 1\% percent of the points being labeled. We developed cross- and intra-sample feature reallocating modules to densely propagate supervision from labeled points to unlabeled points.
We narrowed the performance gap between 3D weakly supervised semantic segmentation and current fully supervised methods while significantly reducing the human effort for annotation. 
 Our proposed method with 10\% and 1\% of the points being labeled can produce compatible results with its fully supervised counterpart in S3DIS and ScanNet dataset.

\section{Acknowledgement}
This research is supported by the Agency for Science, Technology and Research (A*STAR) under its MTC Programmatic Funds (Grant No. M23L7b0021). This research is also supported by A*STAR under its Career Development Award (CDA), Singapore (Award No. 202D8243).
 
 {\small
\bibliographystyle{IEEEtran}
\bibliography{IEEEfull}
}

\begin{IEEEbiography}[{\includegraphics[width=1in,height=1.25in,clip,keepaspectratio]{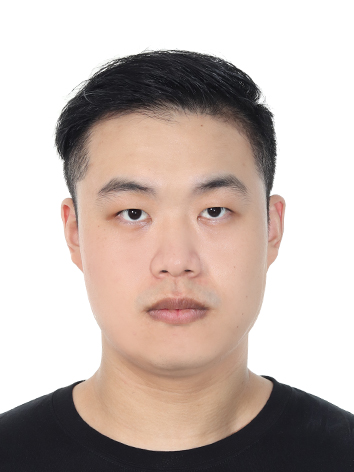}}]{Jiacheng Wei}
received the BEng degree in electronic information engineering from the University
of Electronic Science and Technology of China, and
BEng degree in electronics and electrical engineering
with Honours of the First Class from the University of Glasgow, in 2017. He is currently working
towards the Ph.D. degree with the School of Electrical
and Electronic Engineering, Nanyang Technological
University, Singapore. His research interests include computer vision and machine learning,
with a current focus on 3D vision and generative models.
\end{IEEEbiography}

\begin{IEEEbiography}[{\includegraphics[width=1in,height=1.25in,clip,keepaspectratio]{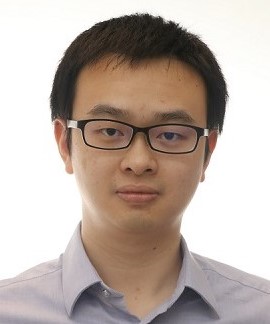}}]{Guosheng Lin}
Guosheng Lin received the bachelor’s and master’s degrees from the South China University of Technology, in 2007 and 2010, respectively, and the PhD degree from
the University of Adelaide, in 2014. He is an assistant professor with the School of Computer Science and Engineering, Nanyang Technological University, Singapore. His research interests include computer vision and machine learning.
\end{IEEEbiography}

\begin{IEEEbiography}[{\includegraphics[width=1in,height=1.25in,clip,keepaspectratio]{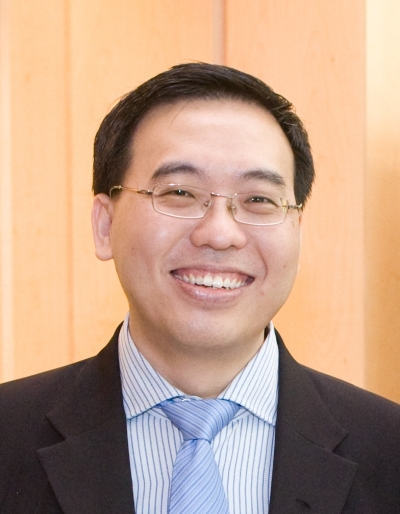}}]{Kim-Hui Yap}
(Senior Member, IEEE) received the bachelor of Electrical Engineering and the Ph.D. degrees both from the University of Sydney, Sydney, NSW, Australia. He is currently an Associate Professor with the School of Electrical and Electronic Engineering, Nanyang Technological University, Singapore. His main research interests include artificial intelligence, data analytics, image/video processing, and computer vision. He has authored more than
100 technical publications in various international peer-reviewed journals, conference proceedings and book chapters. He has also authored a book entitled Adaptive Image Processing:
A Computational Intelligence Perspective, Second Edition published by the CRC Press.

Dr. Yap was an Associate Editor and Editorial Board Member for a number
of international journals. He participated in the organization of various international conferences, serving in different capacities including Technical Program
Co-Chair, Finance Chair, and Publication Chair in these conferences.
\end{IEEEbiography}

\begin{IEEEbiography}[{\includegraphics[width=1in,height=1.25in,clip,keepaspectratio]{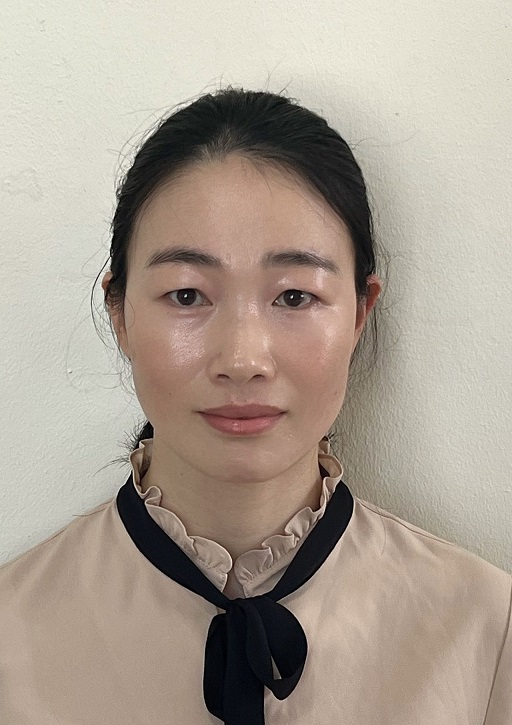}}]{Fayao Liu}
Fayao Liu is a scientist at Institute for Infocomm Research (I2R), A*STAR, Singapore.
She received her PhD in computer science from University of Adelaide, Australia in Dec. 2015.
She mainly works on machine learning and computer vision problems, with particular interests in self-supervised learning, few-shot learning, and generative learning. 
\end{IEEEbiography}

\begin{IEEEbiography}[{\includegraphics[width=1in,height=1.25in,clip,keepaspectratio]{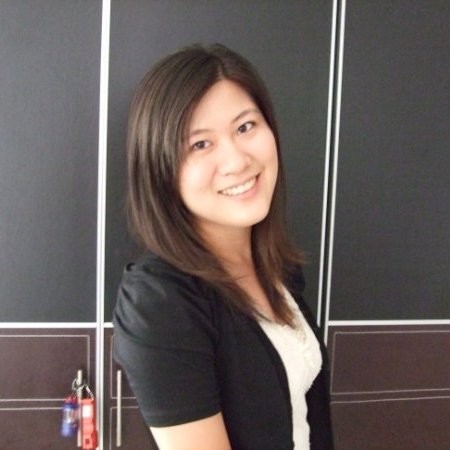}}]{Tzu-Yi Hung}
received the B.Sc. and M.S. degrees from National Chung
Cheng University in 2008 and 2006, respectively, and the Ph.D. degree in
electrical and electronics from Nanyang Technological University in 2015.
She is currently a Data Scientist with the Delta Research Center, Singapore.
She has been a Research Fellow with the Rolls-Royce@NTU Corp Laboratory
since 2014.
\end{IEEEbiography}

\vfill

\end{document}